\documentclass{article}
\usepackage[english]{babel}

\usepackage[preprint]{neurips_2026}
\usepackage[utf8]{inputenc} 
\usepackage[T1]{fontenc}    
\usepackage{hyperref}       
\usepackage{url}            
\usepackage{booktabs}       
\usepackage{amsfonts}       
\usepackage{nicefrac}       
\usepackage{microtype}      
\usepackage{xcolor}         
\usepackage{amsmath}
\usepackage{graphicx}

\title{ChequeMark: An Ensemble Machine Learning Framework for After-Hours Business Deposit Fraud Detection}

\author{
  Ann Youduo Xu\\
  Royal Bank of Canada \\
  \texttt{annxyd12345@outlook.com} \\
  \And
  Emily Yu \\
  Royal Bank of Canada \\
  \texttt{emily.nzyu@gmail.com} \\
  \And
  Justin Leski \\
  Royal Bank of Canada \\
  \texttt{justinleski@icloud.com} \\
  \And
  William Lam \\
  Royal Bank of Canada \\
  \texttt{williamntlam@gmail.com}
}

\begin{document} 
\maketitle

\begin{abstract}
Cheque fraud is a material risk in after-hours business deposit operations because funds may be released within one business day, while cheque clearing takes several days. This timing gap creates a fraud exposure window for financial institutions. Prior mitigation relies on static, deposit-level checks and therefore miss historical client behavior and evolving patterns. To address this gap, we propose a multi-view ensemble ML framework that combines\footnote{This work was completed at Royal Bank of Canada as part of the RBC Amplify program.}: Extreme Gradient Boosting (XGBoost) for known fraud patterns, Isolation Forest for label-free anomaly detection, and Graph Sample and Aggregate (GraphSAGE) for relational patterns associated with transaction activities. We then combine the three outputs into a single client-level risk score. Under stable conditions, performance is comparable to XGBoost; under a targeted distribution shift, our framework performs best (F1: 83.77\%, FPR: 0.69\%) versus XGBoost (F1: 82.77\%, FPR: 0.72\%). These results indicate improved robustness to distribution shift while preserving interpretability through plain-language explanations grounded in behavioral, anomaly, and relational evidence.
\end{abstract}

\section{Introduction}
Commercial banking is a critical sector in banking. Due to higher transaction volumes and more sophisticated transaction behaviors, commercial banking clients present elevated fraud risk, with after-hours business deposit services being one of the exposed workflows. After-hours business deposit services enable commercial banking clients to deposit cheques, coin, and cash outside business hours with next-business-day fund availability. While coin and cash are immediately verifiable upon physical receipt, cheques introduce a distinct fraud vulnerability: a cheque is a deferred payment instrument whose validity cannot be confirmed until it clears with the originating financial institution. While domestic cheques require up to five business days to clear and cheques drawn on foreign financial institutions can take up to thirty days, receiving banks provide next-business-day fund availability to improve customer experience and prevent client attrition to competitors that offer faster access. During this window, the receiving bank bears the full fraud exposure, as funds already withdrawn from a fraudulent cheque cannot be recovered from the originating financial institution.

While the clearing delay creates inherent fraud exposure, several operational factors further compound this risk. First, fraud teams receive only an aggregate deposit total rather than individual cheque details. Second, fraud teams need to manually review each cheque in a fraud alert image by image—a process that can consume up to twenty minutes per alert while drawing on multiple disparate legacy systems. Third, alerts are generated at the transaction level without client context, forcing fraud teams to separately compile client history before they can assess risk.

The intervention opportunity is clear: fraud teams must accurately assess suspicious deposits before clients can withdraw funds. This time-pressured setting makes Machine Learning (ML) valuable because it can integrate client history, deposit behaviors and cheque details, detect complex, non-linear patterns patterns that static rules may miss and assess risks in seconds. Before the adoption of ML methods, fraud detection relied on rules-based assessment systems, where predefined conditional rules determine whether a transaction is flagged as potentially fraudulent. The core limitations of rule-based systems are high false positive rates due to simplistic thresholds, labor-intensive tuning and inability to respond to novel fraud cases.

Recognizing the limitations of rules-based systems, the financial services industry has increasingly turned to ML-based fraud detection. Research in ML spans multiple decades and includes both traditional statistical models and modern deep learning methods. It also includes both supervised learning approaches, which require labels, and unsupervised learning approaches, which identify anomalies without explicit labels.

Despite this rich body of work, existing ML literature for operational fraud detection exhibits four key limitations. First, many prior ML studies \citep{sam02, bhattacharyya11, alwadain23, ajit25} focused on transaction or deposit-level classification, overlooking client behavioral patterns that emerge only when transaction histories are modeled jointly. Second, many of these studies relied on a single model architecture, which can underrepresent complementary fraud evidence, such as relational or anomaly-based signals. In particular, models trained only on tabular transaction features may miss relational signals embedded in counterparty networks. Third, some prior works \citep{whiting12, Hajek17} didn't address operational constraints such as integration into deposit workflows and low inference latency. Fourth, as \citep{Nesvijevskaia21} pointed out, fraud must be explained, especially in the banking sector to protect customers from financial losses, meet legal requirements, and prevent unfair bias. However, many prior ML studies \citep{jurgovsky18, Vanbelle22, Moreira22, Alsuwailem23} provided limited decision-level explainability, which reduces fraud officers' trust and makes regulatory auditability difficult in operational fraud detection.

\subsection{Contribution}

We address these gaps through the following three main contributions.

\paragraph{Client-Level Multi-View Ensemble Framework.} We propose a a client-level, multi-view ensemble ML framework that learns from aggregated client profiles and transaction histories, by ensembling a supervised, an unsupervised, and a graph-based ML model. The framework captures cross-deposit patterns and improves robustness under concept drift.

\paragraph{Operational Deployment in ChequeMark.} We propose and implement deployment of the framework via REST API as a microservice of the fraud detection system we designed. We name the fraud detection system ChequeMark.

\paragraph{Human-Centered Explainability.} We propose a human-centered explainability layer that supports auditability and human-in-the-loop decision making. Explanations are generated in plain language by Large Language Models (LLMs), grounded in SHAP (SHapley Additive exPlanations) feature attributions and gradient-based graph influence and masking.

\subsection{Outline}
Section \ref{sec:literature} reviews work related to our ML framework. Section \ref{sec:problem} formalizes the problem as client-level binary classification with risk scoring objectives and defines our multi-view feature engineering approach. Section \ref{sec:framework} details the proposed multi-ensemble ML framework, training protocols, and explainability mechanisms. Section \ref{sec:experiment} presents experimental results on synthetic after-hours business deposit data. Section \ref{sec:deployment} discusses ML inference microservice deployment considerations. Section \ref{sec:discussion} details the conclusion, limitations and future work.

\section{Related Work}
\label{sec:literature}

Fraud detection literature has extensively applied various single-modality ML methods—approaches that rely on a single algorithm—including supervised learning, unsupervised learning, and graph-based representation learning. We refer to these as \textit{single-learners} to distinguish them from ensemble approaches that combine multiple modalities. While single-learners have demonstrated strong performance in specific contexts, each modality exhibits inherent limitations that constrain detection effectiveness when deployed in isolation. In the following, we first review prior single-learner methods (Section \ref{subsec:singlelearner}), then discuss ensemble approaches (Section \ref{subsec:ensemblelearner}).

\subsection{Single-Learners}
\label{subsec:singlelearner}
\paragraph{Supervised Learning} Supervised learning methods can struggle when fraud pattern drifts and when faced with more complex fraud scenarios such as fraud ring network or laundering behaviors. Probabilistic methods, such as Naïve Bayes classifiers and logistic regression, enable rapid inference and straightforward interpretation of feature coefficients, but they are relatively too simple to effectively detect fraud \citep{bakumenko22, Kumar22}. Distance-based methods, such as Support Vector Machine \citep{kim16} and K-Nearest Neighbors \citep{Alsuwailem23, Moreira22}, adopt hyperplanes and proximity to historical examples to detect fraudulent transactions, but they require prohibitive computational cost during inference on large transaction volumes. In contrast, tree models have a more robust performance. Ensemble trees (Random Forest \citep{Kumar22}) and gradient boosting models (including XGBoost\citep{chen23}, CatBoost\citep{Lokanan22}, LightGBM \citep{fang19}, AdaBoost\citep{Ileberi21}, and RUSBoost\citep{Achakzai22}) consistently outperform across credit card fraud, money laundering, insurance fraud, and payment system anomalies. To address traditional classifiers' inability of catching temporal patterns, some researchers apply Long Short-Term Memory(LSTM) networks to capture transaction sequences \citep{Benchaji21, Esenogho22}. Nevertheless, these algorithms all fail to adapt to unseen fraud tactics without retraining. 

\paragraph{Unsupervised Learning} Unsupervised learning methods complement the limitation of supervised methods by its adaptability to evolving fraud tactics. As fraudsters develop novel schemes, unsupervised models can continue identifying statistical deviations without requiring labeled retraining data. Density-based, distance-based and tree-based unsupervised methods have proven effective for real-time fraud or anomaly monitoring. \citep{Vanhoeyveld20} applied Fixed Width Anomaly Detection, Local Outlier Factor (LOF), and boxplot-based techniques to Value Added Tax declarations, demonstrating high predictive power for fraud detection. Later, \citep{Domashova22} analyzed Density-Based Spatial Clustering of Applications with Noise (DBSCAN), LOF, Isolation Forest, and Elliptic Envelope to identify atypical bank transactions, revealing that unsupervised anomaly detection captures contextual irregularities missed by supervised methods. Another effective algorithm is Autoencoders, the resconstruction-based deep learning method that mark anomalies by learning from compressed representations of normal patterns \citep{Misra20}. Variational Autoencoders (VAE) extend the capability to sophisticated minority fraud synthesis that can address class imbalance in realistic fraud scenarios \citep{Tingfei20}. The problem with unsupervised methods, however, is an elevated false-positive rate without labeled guidance, which can harm operational performance of financial institutions.

\paragraph{Graph Representation Learning} Graph representational learning techniques, with particular attention on learning node embeddings, add another pieces to the puzzle by constructing a fraud graph that maps networks of receivers and senders. One of the most important contributions was Graph Convolutional Networks (GCN)\citep{defferrard17} which rely heavily on node features and support supervised learning on large and sparse graphs. GCN applications in fraud detection settings include \citep{liu20}, \citep{usman23}. However, GCNs are largely transductive approaches that require the entire graph during node embedding generation, forcing continuous retraining every time new transactions come in. As a result, most cutting-edge research turned to examine the inductive learnings—for example Graph Sample and Aggregate (GraphSAGE) \citep{hamilton18}—that can handle unseen nodes and enable real-time inference for high-throughput fraud detection. Building on this, \citep{liu21} integrated an attention mechanism into GraphSAGE that reduces interference from abnormal samples while amplifying influence of normal ones. \citep{Vanbelle22} also tested the performance of GraphSAGE and Fast Inductive Graph Representation Learning (FI-GRL) on highly imbalanced credit card transaction networks. But a key drawback is: node embeddings learned via message passing are encoded and decoded through multiple non-linear transformations, making them difficult to interpret. Unlike tabular features with direct business meaning, embeddings lack transparency and may lose domain-specific information during compression.

\subsection{Ensemble-Learners}
\label{subsec:ensemblelearner}
This heterogeneity in single-learners' strengths and weaknesses has motivated us to explore ensemble frameworks that combine complementary algorithmic perspectives. The fundamental principle of ensemble-learners are consistent: to add robustness, reduce variance and bridge the performance gap of any individual constituent algorithm, though the mechanisms for aggregation are diverse.

Simple ensemble frameworks include hard voting, soft voting and weighted averaging. Hard voting integrates predictions by summing votes across models, with highest-voted class declared winner \citep{baker22}. Soft voting alternatively averages class probability estimates across models for more refined aggregation. Weighted averaging is another way to optimize combination weights based on model performance. \citep{shou23} demonstrated this approach on financial statement fraud, developing a weighted ensemble combining Naive Bayes and K-Nearest Neighbors with a 2:1 weight ratio after multi-metric evaluation. 

Learned ensemble frameworks, on the other hand, train secondary models to discover optimal combinations of base learner predictions. For example, \citep{Achakzai22} employed meta-learners that learn weighted combinations of base model outputs and capture non-obvious synergies and interaction effects between constituent models for Chinese financial statement fraud detection. \citep{Chullamonthon23} ensembled the results of a supervised LSTM and unsupervised LSTM-AutoEncoder, which are then fed into an SVM classifier that learns the optimal way to combine them. The results often outperform constituent algorithms.

\section{Problem Formulation}
\label{sec:problem}

In this section, we describe the underlying probabilistic learning problem that formalizes the fraud detection task, while stating our proposed multi-view ensemble ML framework as an estimator for this problem. 

\paragraph{Random Variables and Notations.} Let $X \in \mathbb{R}^d$ denote the tabular features of a client, $G$ denote the graph encoding relationships among entities, and $Y \in \{0,1\}$ denote the fraud label. Assume the observations are drawn from an unknown joint distribution $(X, G, Y) \sim P(X, G, Y)$. The goal is to estimate the conditional probability $P(Y=1|X,G)$, which is unknown.

\paragraph{Decision Rule And The Difficulty.} We define a classifier $f: \mathcal{X} \times \mathcal{G} \rightarrow [0,1]$ where $f(X,G)$ maps a feature instance $X$ and a graph $G$ to a probability score in $[0,1]$, estimating the conditional probability $P(Y=1|X,G)$. 
The final binary prediction $\hat{Y} \in \{0,1\}$ is determined by a hard decision threshold $\tau \in [0,1]$, $\hat{Y} = 1$ if $f(X,G) > \tau$, otherwise $\hat{Y} = 0$. 
The optimal classifier is $f^*(X,G) = P(Y=1|X,G)$, but estimating this distribution is difficult because: 
in practice, (1) only a subset of frauds are labeled due to reliance on manual human reporting, delayed discovery, and extreme data volume, 
(2) fraud tactics evolve over time because fraudsters continuously adapt their strategies, (3) graph dependence violates the assumption that observations are independent,
and (4) some novel frauds correspond to low-density regions of the feature space, making them hard to distinguish from rare legitimate behaviour. 
Consequently, $P(Y|X,G)$ cannot be accurately estimated from labeled tabular data alone.

\paragraph{Multi-View Decomposition.} Instead of approximating $P(Y|X,G)$ using a single model, we decompose the evidence into three complementary components:

\textit{Predictive Evidence (Labeled Covariates):} We estimate a calibrated score $f_{\text{pred}}(X) \approx P(Y=1|X)$ from labeled data. This component captures direct associations between client-level covariates and fraud risk.

\textit{Anomaly Evidence (Distributional Deviation):} We estimate an anomaly score $f_{\text{anom}}(X) = A(X)$, which increases as an observation becomes less typical under the marginal data distribution $p(X)$. I.e., $A(X)$ corresponds to inverse association with $p(X)$, which helps surface novel or weakly labeled fraud patterns.

\textit{Relational Evidence (Graph Dependence):} We estimate a structural risk score $f_{\text{rel}}(X,G)$ using neighborhood-dependent information derived from the interaction graph. This component captures dependence patterns (e.g., local connectivity, flow concentration, and relational context) that are not identifiable from independent tabular rows.

Aggregating these evidence, we estimate the conditional probability $P(Y=1|X,G)$ using a straightforward weighted linear combination $F(f_{\text{pred}}, f_{\text{anom}}, f_{\text{rel}})$ to preserve model interpretability:
\begin{align}
F(f_{\text{pred}}, f_{\text{anom}}, f_{\text{rel}}) = w_1 \cdot f_{\text{pred}} + w_2 \cdot f_{\text{anom}} + w_3 \cdot f_{\text{rel}}, \quad \text{where} \quad \sum_{i=1}^{3} w_i = 1
\end{align}

\paragraph{Learning Objective.} The ultimate learning objective can be written as:
\begin{align}
\min_{f} \mathbb{E}_{(X,G,Y) \sim P}[\mathcal{L}(Y, f(X,G))]
\end{align}
where $\mathcal{L}$ is a classification loss (e.g., cross-entropy). Since $f$ is difficult to estimate directly, we approximate it by combining three estimators:
\begin{align}
f(X,G) = F(f_{\text{pred}}(X), f_{\text{anom}}(X), f_{\text{rel}}(X,G))
\end{align}

\paragraph{Explainability objective.} To add explainability, we formalize this as a post-hoc attribution problem: given a trained classifier $f(X,G)$ and an instance $(X_i, G_i)$ with prediction $\hat{y}_i = f(X_i, G_i)$, we seek to identify (1) a set of salient features $S \subseteq \{1, \ldots, d\}$ that most strongly contribute to $\hat{y}_i$ in the tabular components, and (2) a set of influential graph neighbors $N_i \subseteq V(G)$ that propagate risk through the network structure in the graph component. The attribution mechanism must satisfy interpretability constraints: feature contributions should be quantifiable, neighbor influences should be traceable, and the combined explanation should be translatable into plain language for fraud officers without ML expertise.

\section{Proposed Framework}
\label{sec:framework}

\subsection{Constituent Algorithms} 

\paragraph{XGBoost as Supervised Fraud Detection Model.}

We select Extreme Gradient Boosting (XGBoost) based on its ability to capture complex non-linear feature interactions through gradient boosting. Its built-in L1/L2 regularization also prevents overfitting on the minority class. XGBoost is a gradient boosting framework that builds an ensemble of decision trees sequentially, where each tree corrects errors of previous trees. Given client features $X \in \mathbb{R}^d$, the predictive component estimates fraud probability~\citep{Chen16}:
\begin{align}
f_{\text{pred}}(X) = \sigma\left(\sum_{k=1}^{K} f_k(X)\right)
\end{align}
where $f_k$ represents the $k$-th decision tree, $K$ denotes the total number of boosting rounds, and $\sigma(\cdot)$ is the logistic sigmoid function mapping scores to probabilities. The model minimizes regularized binary cross-entropy loss with class weighting to address class imbalance~\citep{Chen16}:
\begin{align}
\mathcal{L}_{\text{pred}} = -\frac{1}{N}\sum_{i=1}^{N} v_i \left[Y_i \log f_{\text{pred}}(X_i) + (1-Y_i)\log(1-f_{\text{pred}}(X_i))\right] + \sum_{k=1}^{K}\Omega(f_k)
\end{align}
where $N$ is the number of training samples, $v_i$ assigns higher weight to minority class samples (fraud cases), and $\Omega(\cdot)$ is a regularization term that penalizes tree complexity (depth, number of leaves) across the $K$ boosting rounds to prevent overfitting.

\paragraph{Isolation Forest as Unsupervised Anomaly Detection Model.}

We select Isolation Forest because of linear time complexity, a low memory requirement and the capacity to scale up to handle extremely large data size and high-dimensional problems~\citep{liu08}. Isolation Forest detects anomalies by measuring how easily a sample can be isolated in feature space. The algorithm builds an ensemble of isolation trees, where each tree recursively partitions the feature space via random splits, and anomalous points require fewer splits to isolate. For client features $X \in \mathbb{R}^d$, the anomaly score is computed as~\citep{liu08}:
\begin{align}
f_{\text{anom}}(X) = 2^{-\frac{E[h(X)]}{c(n)}}
\end{align}
where $h(X)$ represents the path length required to isolate $X$ in a random tree, $E[\cdot]$ denotes the expectation over all trees in the ensemble, and $c(n)$ is the average path length normalization factor for a dataset of size $n$. Scores closer to 1 indicate anomalies (short isolation paths), while scores near 0 indicate normal instances (long isolation paths).

\paragraph{GraphSAGE as Graph Representation Learning Model.}

We select GraphSAGE in order to identify network behaviors after analyzing multiple transductive and inductive Graph Neural Network (GNN) methods on the transaction network. GraphSAGE learns node embeddings by aggregating features from local neighborhoods via message passing, enabling detection of fraud rings and coordinated schemes invisible to tabular models. For the transaction graph $G = (\mathcal{V}, \mathcal{E})$, we employ a multi-layer GraphSAGE architecture with hybrid edge aggregation. At layer $\ell$, node $v$'s embedding is updated via~\citep{hamilton18}:
\begin{align}
\mathbf{h}_v^{(\ell)} = \text{ReLU}\left(\mathbf{W}^{(\ell)} \cdot \text{CONCAT}\left(\mathbf{h}_v^{(\ell-1)}, \text{AGG}^{(\ell)}\left(\{\mathbf{h}_u^{(\ell-1)} : u \in \mathcal{N}(v)\}\right)\right)\right)
\end{align}
where $\mathcal{N}(v)$ denotes the neighborhood of node $v$, $\text{AGG}^{(\ell)}$ is the aggregation function, $\mathbf{W}^{(\ell)}$ is a learnable weight matrix, $\text{ReLU}(\cdot)$ is the Rectified Linear Unit activation function, and $\text{CONCAT}(\cdot, \cdot)$ denotes vector concatenation. We employ a hybrid aggregator combining max, mean, and sum pooling to capture diverse neighborhood patterns:
\begin{align}
\text{AGG}_{\text{hybrid}}(\{\mathbf{h}_u^{(\ell-1)}\}) = \text{CONCAT}\left(\max_u \mathbf{h}_u^{(\ell-1)}, \frac{1}{|\mathcal{N}(v)|}\sum_u \mathbf{h}_u^{(\ell-1)}, \sum_u \mathbf{h}_u^{(\ell-1)}\right)
\end{align}
This hybrid framework captures local extremes (max), average patterns (mean), and aggregate signals (sum), enabling the model to identify both individual high-risk neighbors and collective neighborhood characteristics indicative of fraud networks. The final relational component $f_{\text{rel}}(X, G)$ is obtained by passing the learned node embeddings through a classification head that outputs fraud probability.

\subsection{Multi-view Ensemble ML Framework}

\begin{figure}[ht]
    \centering
    \includegraphics[width=0.5\linewidth]{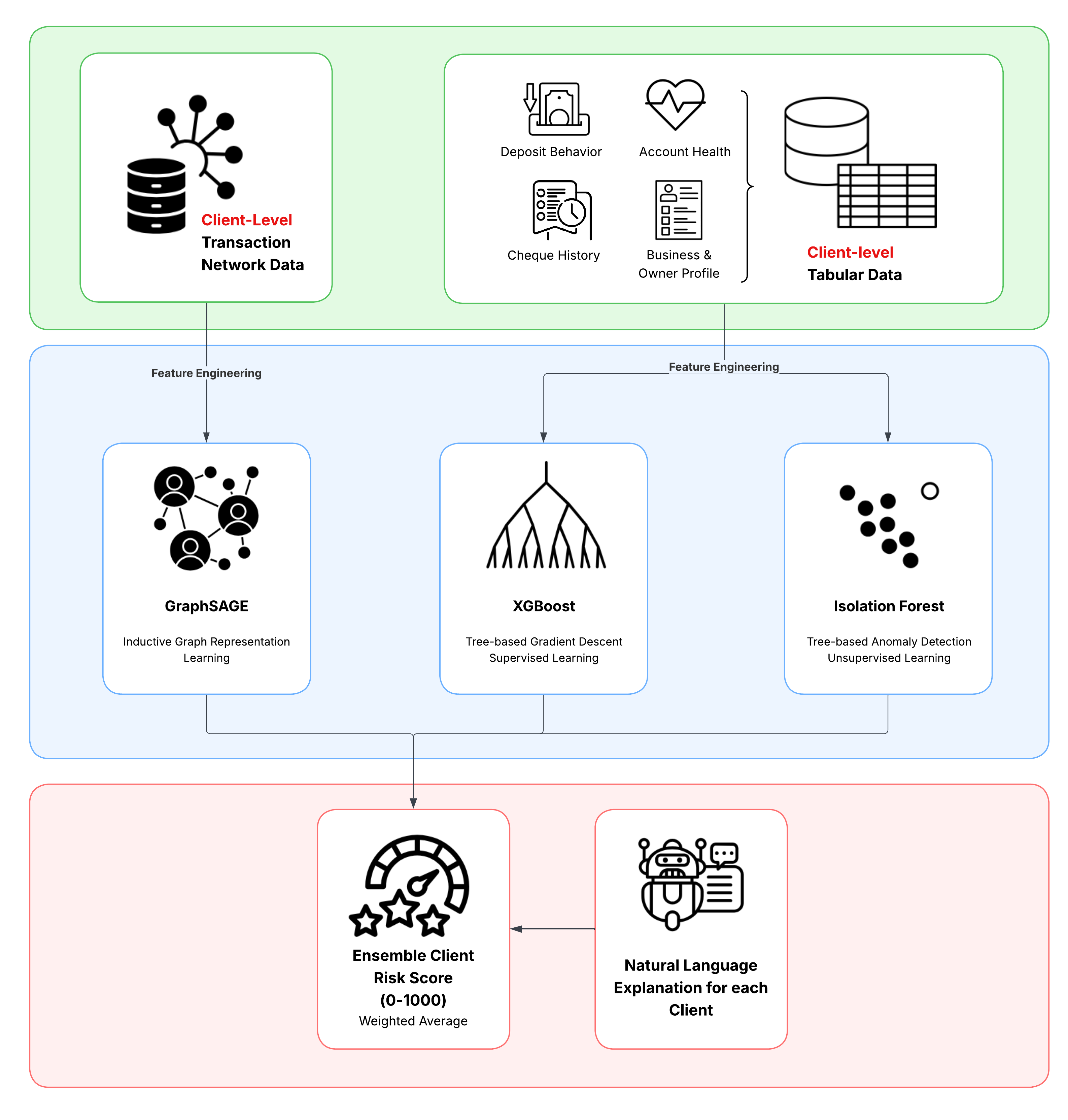}
    \caption{Multi-view Ensemble ML Framework}
    \begin{center}
        \small
        The framework consists of three layers: (1) a data ingestion layer that aggregates raw transaction records into client-level transaction networks and tabular features, (2) a model layer where three constituent algorithms process their respective inputs independently, and (3) an aggregation layer that combines model outputs via weighted averaging and generates explainable rationales for fraud officers.
    \end{center}
    \label{fig:ml_architecture}
\end{figure}

Our multi-view ensemble ML framework (see Figure \ref{fig:ml_architecture}) addresses fundamental limitations of single-learner approaches: supervised models excel at known patterns but fail against novel schemes; unsupervised models adapt to evolving tactics but generate high false-positive rates; graph models capture network structures but cannot leverage tabular features. By systematically combining these modalities, the ensemble achieves robustness against concept drift while maintaining high precision and recall.

Following the multi-view decomposition in Section \ref{sec:problem}, we implement the three complementary components in the model layer:

1. \textbf{$f_{\text{pred}}(X)$}: XGBoost gradient boosting on tabular features $X$, optimizing binary cross-entropy with class weight adjustment;
2. \textbf{$f_{\text{anom}}(X)$}: Isolation Forest tree-based unsupervised anomaly detection on $X$;
3. \textbf{$f_{\text{rel}}(X, G)$}: GraphSAGE graph neural network on transaction graph $G$, learning node embeddings via 2-hop message passing and predicting fraud via supervised classification on labeled nodes.

Each model outputs a fraud probability or anomaly score normalized to $[0,1]$. The final client risk score is computed via the weighted linear combination defined in Section \ref{sec:problem} where weights are determined via PR-AUC (Precision-Recall AUC) proportional allocation, with each model's weight proportional to its individual test set PR-AUC:
\begin{align}
w_i = \frac{\text{PR-AUC}_i}{\text{PR-AUC}_{\text{pred}} + \text{PR-AUC}_{\text{anom}} + \text{PR-AUC}_{\text{rel}}}, \quad \sum_{i=1}^{3} w_i = 1
\end{align}
Clients are flagged for investigation if $f(X, G) \geq \tau_{\text{op}}$, with threshold $\tau_{\text{op}}$ set based on operational constraints.

\subsection{Training and Ensemble Weight Optimization}

Our model development protocol consists of two stages: (1) independent training of each constituent model on the same stratified train-test split, and (2) ensemble weight determination via PR-AUC proportional allocation. All models are trained on a deterministic client-level stratified split. Stratification maintains the same fraud prevalence ratio in both partitions to ensure representative evaluation.

Each constituent model is trained independently through cross-validation. $f_{\text{pred}}$ (XGBoost) is trained via gradient boosting with binary cross-entropy loss and class weighting to address imbalance, using early stopping to prevent overfitting. $f_{\text{anom}}$ (Isolation Forest) is trained without labels by building an ensemble of isolation trees on random subsamples. $f_{\text{rel}}$ (GraphSAGE) is trained via supervised node classification with binary cross-entropy loss on labeled training nodes, allowing message passing over the full graph while restricting supervision to training labels. After independent training, ensemble weights are determined via PR-AUC proportional allocation, where each model's weight is proportional to its individual test set PR-AUC. All constituent scores are normalized to a common range before weighted aggregation into the final ensemble score $f(X, G)$.

\subsection{Explainability Mechanisms}

Operational fraud detection requires transparency at multiple levels of abstraction to serve different stakeholders and decision contexts. Thus, we employ three complementary explainability mechanisms: (1) SHAP values provide rigorous feature attributions for models trained on tabular dataset; (2) gradient-based counterfactual graph analysis that identify which transaction counterparties propagate fraud risk through the graph structure; and (3) LLM-based plain language translates numbers into business-focused narratives for fraud officers without ML expertise. These three mechanisms work synergistically.

\paragraph{SHAP Values}

For tree-based $f_{\text{pred}}$ and $f_{\text{anom}}$, we employ SHAP TreeExplainer, a model-agnostic explanation framework grounded in cooperative game theory. SHAP values quantify each feature's contribution to the predicted client risk score \citep{lundberg17}. Positive SHAP values indicate that the feature increases fraud risk, while negative values indicate that the feature decreases risk, with magnitude reflecting contribution strength. TreeExplainer leverages the tree structure to compute exact SHAP values efficiently, enabling real-time explanations during inference.

\paragraph{Graph Influence Analysis} 

The graph influence analysis method is built on the method introduced by \citep{ying19}. For $f_{\text{rel}}(X, G)$, we identify influential neighbors via gradient-based analysis to explain graph-based predictions and decode the learned node embeddings. We compute neighbor influence as:
\begin{align}
I(u \to v) = \left|\frac{\partial P_{\text{rel}}(y_v=1)}{\partial \mathbf{h}_u}\right|
\end{align}
where $I(u \to v)$ measures how much neighbor $u$'s embedding $\mathbf{h}_u$ affects node $v$'s fraud probability $P_{\text{rel}}(y_v=1)$ through backpropagation, with higher values indicating stronger influence. This gradient-based approach identifies which transaction partners most strongly influence a client's risk score, revealing potential fraud rings. We complement this with counterfactual influence via neighbor masking:
\begin{align}
\Delta P(u \to v) = P_{\text{rel}}(y_v=1 \mid \mathcal{N}(v)) - P_{\text{rel}}(y_v=1 \mid \mathcal{N}(v) \setminus \{u\})
\end{align}
where $\mathcal{N}(v)$ is the full neighborhood of $v$ and $\mathcal{N}(v) \setminus \{u\}$ is the neighborhood with $u$ removed. This counterfactual $\Delta P(u \to v)$ quantifies the change in fraud probability when removing neighbor $u$ from the graph, providing causal explanations for fraud officers to understand network-based risk propagation.

\paragraph{LLM-Based Plain Language Explanations}

After extracting SHAP values and graph influence scores, these technical attributions are translated into plain-language rationales via Large Language Model (LLM) integration. This bridges the gap between model outputs and human understanding. The microservice will prompt an LLM (e.g. GPT-4) with structured JSON inputs containing feature importance vectors and neighbor influence scores, instructing it to generate concise bullet points (typically 2-6) ordered by impact while avoiding statistical jargon and technical terminology. The LLM-based approach is preferred over template-based explanations because it adapts phrasing to the specific fraud pattern.

\section{Experimentation Results}
\label{sec:experiment}

\subsection{Experimental Setting}

\paragraph{Data.}
The dataset employed in this study is fully synthetic and generated via a rule-based, probabilistic simulation pipeline designed to replicate operational characteristics of after-hours business deposit fraud detection. The synthetic generation process produces four interconnected data sources: (1) after-hours business deposit transaction records, (2) cheque-level records, (3) business client profiles, and (4) account transaction history. The dataset comprises 17,093 after-hours business deposit clients, 480,000 deposit slips, 2,000,000 cheques, and 1,500,000 account transactions spanning 12 months. From these raw data sources, we engineer 101 tabular features per client and a transaction network with 25,093 nodes and 967,522 directed edges. While synthetic data enables controlled experimentation and preserves privacy, tree-based models can directly learn the underlying generation rules, which may inflate reported performance metrics. Consequently, absolute performance metrics reported herein should be interpreted as upper bounds; final threshold tuning and policy decisions require validation against governed real data. 

Detailed data generation methodology, schema validation, and feature engineering procedures are provided in Appendix \ref{appendix:data}.

\paragraph{Training and Testing}
The three constituent models are trained independently on a training set of 13,675 after-hours business deposit clients (350 fraud cases, 2.57\% prevalence). A held-out test set of 3,418 clients (88 fraud cases, 2.57\% prevalence) is used for model evaluation. The train-test split is performed at the client level using stratified sampling to maintain fraud prevalence ratios. Since client are strictly separated, this splitting protocol doesn't introduce data leakage. However, it does not reflect realistic scenarios where models are trained on all historical data up to a cutoff date and evaluated on future fraud cases. A sequential train-testsplit where all training fraud cases precede all test cases in calendar time would provide more rigorous evaluation. Ensemble weights are optimized on the stable distribution test set using PR-AUC proportional allocation, and these fixed weights are applied without retuning to another distribution-shifted test set to enable evaluation of robustness to concept drift.

\paragraph{Hyperparameters}
Hyperparameters are selected via Optuna optimization with 5-fold stratified cross-validation maximizing PR-AUC. We conduct 50 trials with MedianPruner for XGBoost and Isolation Forest, and 15 trials for GraphSAGE due to higher computational cost. The final hyperparameter configurations for each model are provided in Appendix \ref{appendix:hyperparameters}.

\subsection{Experimental Results}

We evaluate the proposed framework on two experimental settings to assess both baseline performance and robustness of our framework to distribution shift. First, we report performance on the stable test distribution where fraud patterns follow the same distribution as training data. Under stable conditions, we expect the ensemble to perform comparably to its strongest constituent model (i.e. XGBoost), as the supervised component dominates when fraud patterns are well-represented in training data. Second, we evaluate on a distribution-shifted test set that simulates emerging fraud scenarios where fraudsters exploit newly opened accounts that are underrepresented in training data. The shifted test set is constructed by filtering clients with account age less than 18 months. Under distribution shift, we expect the ensemble to outperform individual constituents through complementary signal aggregation, as unsupervised and graph-based components compensate when supervised learning fails to generalize. 

Our results confirm these 2 hypotheses.
 
\paragraph{Performance on Stable Distribution}

Table \ref{tab:stable_performance} reports the performance of the full ensemble and its constituent models on the full test set at a 3\% operational flagging threshold. The ensemble aggregates the three models via PR-AUC proportional weighting. The full ensemble achieves Precision 0.8058, Recall 0.9432, and F1 0.8586, capturing 83 of 88 fraud cases while maintaining a low false positive rate of 0.63\%. As expected, the ensemble performance closely tracks XGBoost ($f_{\text{pred}}$), which achieves nearly identical metrics, reflecting the supervised component's dominance when fraud patterns are well-represented in training data.

\begin{table}[h]
\centering
\caption{Performance on Stable Test Distribution}

\label{tab:stable_performance}
\begin{tabular}{lcccccc}
\hline
\textbf{Model} & \textbf{Precision} & \textbf{Recall} & \textbf{F1} & \textbf{FPR} \\
\hline
Proposed Method (Full Ensemble) & 0.8058 & 0.9432 & 0.8586 & 0.0063 \\
Predictive Component ($f_{\text{pred}}$)  & 0.8058 & 0.9432 & 0.8691 & 0.0060 \\
Anomaly Component ($f_{\text{anom}}$) & 0.1359 & 0.1591 & 0.1466 & 0.0267 \\
Relational Component ($f_{\text{rel}}$) & 0.1456 & 0.1724 & 0.1579 & 0.0264 \\
\hline
\end{tabular}
\end{table}

\paragraph{Performance Under Distribution Shift}

However, the ensemble's primary advantage lies not in marginal gains under stable conditions, but in its robustness to shifting fraud landscapes. To evaluate robustness against concept drift and novel fraud patterns, we conduct a stress test on a distribution-shifted test subset where 20.7\% of clients have $\text{account\_age\_months} < 18$, compared to the natural 5.6\% in a random stratified split. The 18-month threshold is selected based on operational domain knowledge and typical fraud incubation periods observed in banking fraud literature, where first-party fraudsters often nurture healthy credit profiles for over months or years before bust-out \citep{bai2013, experian2024, earlywarning2024}. Fraud prevalence still remains balanced at 2.57\% overall. This simulates the operational scenario where models trained on established client histories must generalize to cold-start risk assessment for early-tenure clients with sparse behavioral context, higher onboarding volatility, and elevated fraud risk. Table \ref{tab:drift_performance} reports the full ensemble's and constituent models' performance on the distribution-shifted subset at the 3\% operational threshold.

\begin{table}[h]
\centering
\caption{Performance on Shifted Test Distribution}
\label{tab:drift_performance}
\begin{tabular}{lcccccc}
\hline
\textbf{Model} & \textbf{Precision} & \textbf{Recall} & \textbf{F1} & \textbf{FPR} \\
\hline
\textbf{Proposed Method (Full Ensemble)} & \textbf{0.7767} & \textbf{0.9091} & \textbf{0.8377} & \textbf{0.0069} \\
Predictive Component ($f_{\text{pred}}$)  & 0.7670 & 0.8977 & 0.8272 & 0.0072 \\
Anomaly Component ($f_{\text{anom}}$) & 0.2039 & 0.2386 & 0.2199 & 0.0246\\
Relational Component ($f_{\text{rel}}$) & 0.1553 & 0.1818 & 0.1675 & 0.0261\\
\hline
\end{tabular}
\end{table}

The results empirically confirms the ensemble design's rationale: by combining supervised, unsupervised, and graph-based models, the framework achieves graceful degradation under distribution shift. When $f_{\text{pred}}$ fails to generalize to novel fraud patterns absent from training data, $f_{\text{anom}}$ and $f_{\text{rel}}$ compensate by detecting outliers and network anomalies that do not require labeled historical precedent. This multi-view robustness is essential for operational fraud detection, where fraudsters continuously adapt tactics to evade detection, and model retraining lags behind emerging threats.

\section{Deployment and Operations}
\label{sec:deployment}

This section describes the operational deployment of our proposed multi-view ensemble ML framework, which we refer to as the ML inference microservice. The ML inference microservice operates within the broader ChequeMark fraud detection system alongside other frontend and backend components that integrates into after-hours business deposit processing workflows. Subsection \ref{subsec: systemarchitecture} describes the currently deployed microservice architecture. Subsection \ref{subsec:error} presents operational performance metrics validated through local load testing and API latency measurements. Subsection \ref{subsec:retraining} outlines a proposed retraining pipeline that is not yet deployed because automated retraining requires access to real, production data with confirmed fraud labels for model tuning and validation.

\subsection{ML Inference Microservice Architecture}
\label{subsec: systemarchitecture}

The ML inference microservice operates as part of the ChequeMark backend architecture that targets completion within the time-constrained processing window for a full day's after-hours business deposits.

The microservice follows a two-tier design: (1) a synchronous ML inference API delivering risk scores and (2) an asynchronous explanation worker that polls the database every 10 seconds to generate SHAP values, graph influence scores, and LLM-based plain-language rationales. This decoupling ensures that hold decisions are not delayed by computationally expensive explainability computations. The ML inference API is containerized with Docker and deployed as a FastAPI microservice on a Kubernetes-based orchestration platform, exposing REST endpoints for batch client risk scoring.

\textbf{Validated Performance Metrics.} Performance characteristics are validated through Quality Assurance Testing (QAT) and User Acceptance Testing (UAT) environments testing suites integrated into the Continuous Integration and Continuous Deployment (CI/CD) pipeline and local load testing on representative batch sizes (100-500 clients per request). Service Level Agreements (SLAs) define maximum acceptable response times for operational acceptance in QAT and UAT. Documented performance metrics include: (1) \textit{End-to-end batch latency}: 2-3 seconds per batch for typical workloads, including database I/O, feature engineering, and parallel three-model inference; (2) \textit{Concurrent request handling}: up to 50 concurrent batch requests via semaphore-based concurrency control, preventing out-of-memory failures on 16GB RAM pods; (3) \textit{Cache fallback performance}: distributed caching for client profiles and transaction history features with automatic database fallback when cache is unavailable, adding an estimated 1-2 seconds to batch latency. The microservice meets the 10-second timeout requirement imposed by the upstream backend service for ML inference API. Asynchronous explanation generation is excluded from this latency measurement and timeout requirement, as it occurs post-scoring that does not block hold decisions. Table~\ref{tab:latency} summarizes the latency breakdown for the ML inference microservice.

\begin{table}[h]
\centering
\caption{ML Inference Microservice Latency Breakdown}
\label{tab:latency}
\begin{tabular}{lcc}
\hline
\toprule
\textbf{Component} & \textbf{Latency}\\
\hline
\midrule
\textbf{ML Inference API Batch Scoring} & & \\
\quad End-to-end latency per batch & 2-3s\\
\quad \quad - Database I/O & Included\\
\quad \quad - Feature engineering & Included\\
\quad \quad - Parallel model inference & Included\\
\midrule
\textbf{Cache Performance} & & \\
\quad Cache hit & \textasciitilde50-200ms\\
\quad Cache miss (fallback) & +1-2s\\
\midrule
\textbf{Operational SLAs} & & \\
\quad timeout & 10s\\
\quad QAT SLA (single client) & \textless 10s\\
\quad QAT SLA (batch) & \textless 30s\\
\quad UAT SLA (batch) & \textless 60s\\
\midrule
\textbf{Concurrency} & & \\
\quad Max concurrent requests & 50\\
\quad Memory limit per pod & 16GB\\
\quad Chunk size & 1,000 clients\\
\bottomrule
\hline
\end{tabular}
\end{table}

\textbf{Scalability Strategy.} Scalability to larger production volumes can be achieved through horizontal scaling. It can accommodate larger production volumes through multiple FastAPI pods behind a load balancer, with each pod handling independent batch requests subject to the 50-request semaphore limit. Database query optimization via connection pooling, prepared statements, and read replicas provides additional headroom. For the current scale of synthetic data, the in-memory Pandas-based pipeline is maintainable and cost-effective. However, if daily scoring volume exceeds 1 million clients/day or training data grows beyond 100 million transactions, we recommend considering migration to Spark-based data processing. For graph scalability beyond current volumes, we recommend adopting a graph database (e.g., Neo4j) for efficient subgraph sampling and neighbor retrieval, implementing mini-batch training and inference to bound memory usage, and exploring inductive GraphSAGE variants that compute embeddings without full-graph message passing. These are proposed enhancements, not currently implemented.
 
\subsection{Error Handling and Recovery}
\label{subsec:error}
 
The microservice implements multi-tier error handling to ensure operational resilience, validated through fault injection testing during development. Graceful degradation handles component failures by continuing with reduced functionality: if cache is unavailable, it falls back to database queries; if historical data is missing for a client, features are zero-filled and scoring proceeds with default values; if feature importance analysis fails, it returns scores without feature importance metadata (logged but non-blocking); if LLM integration times out, structured JSON explanations are returned without plain-language synthesis. These fallbacks ensure that fraud officers receive risk scores even under partial microservice failures, prioritizing availability over completeness.
 
Hard failures that prevent scoring—such as model loading errors, database constraint violations, or inference exceptions—mark the job as failed and return HTTP 500 errors. Failed jobs are automatically retried via a background retry worker that polls every 60 seconds, implementing exponential backoff for up to 3 retry attempts. Persistent failures after 3 attempts are marked as permanently failed and trigger operational alerts.
 
The microservice also implements checkpoint-based recovery that enables resumption from partial failures. The scoring pipeline persists risk score records and last-processed chunk metadata after each 1,000-client chunk. If a batch job crashes at chunk 8 of 10, the retry worker resumes processing from chunk 9, avoiding redundant computation. Similarly, the explanation worker tracks explanation status per client (\texttt{pending}, \texttt{completed}, \texttt{failed}) and retries failed explanations up to 3 times before marking them as permanently failed. This granular retry mechanism ensures that transient failures do not require full batch reprocessing.

These design parameters are chosen based on typical transient failure recovery times observed during development to balance retry latency and database load.
 
\subsection{Model Retraining and Monitoring}
\label{subsec:retraining}
 
Model retraining and versioning pipelines are not deployed yet. However, we suggest that model retraining should follow a semi-automated pipeline informed by regulatory approval processes in financial institutions. It should be noted that frequent retraining risks performance degradation from accumulated concept shift.
 
We also suggest implementing a continuous monitoring pipeline. When performance degrades below thresholds (e.g., 30-day average PR-AUC drops $>5\%$ relative to baseline) or when significant distribution shifts are detected (e.g., Kolmogorov-Smirnov test $p < 0.01$ for key features), the monitoring pipeline would trigger retraining alerts. Upon approval from relevant stakeholders, the retraining pipeline would extract 12 months of labeled data, retrain all three constituent models with Optuna hyperparameter tuning, and test performance on a held-out test set. Ensemble weights would be recalibrated via PR-AUC proportional allocation on the new validation set.
 
In addition, as part of the monitoring pipeline, we suggest that model versioning could be managed through a centralized JSON registry that stores version number, training date, hyperparameters, ensemble weights, performance metrics, and deployment status (\texttt{shadow}, \texttt{production}, \texttt{retired}) for each model iteration. Prediction logs would include model version fields to enable retrospective analysis of which model version produced each historical score.

\section{Conclusion}
\label{sec:discussion}

We present a multi-view ensemble ML framework for after-hours business deposit fraud detection that addresses critical gaps in operational fraud detection. The framework combines XGBoost for supervised behavioral prediction, Isolation Forest for unsupervised anomaly detection, and GraphSAGE for graph-based network analysis. On held-out test data with stable distribution, the proposed framework achieves performance comparable to XGBoost alone. On held-out test data with shifted distribution simulating concept drift, the proposed framework achieves greater, more robust performance of 83.73\% F1-score and 0.69\% false positive rate, higher than any of its constituent model. The framework's primary contributions are its ability to capture comprehensive fraud evidence across multiple views and its robustness to distribution shift or evolving fraud patterns.

Beyond predictive robustness, the framework addresses operational requirements through three design principles. First, it introduces client-level risk scoring that evaluates deposits using each client's full behavioral history rather than isolated deposit-level rules. This contribution helps fraud officers make more context-aware decisions and reduce avoidable false positives from one-size-fits-all alert logic. Second, the ML inference microservice, built on this framework, targets low latency and integration with existing deposit workflows. This contribution helps ensure timely fraud interception. Yet full production latency and reliability should be validated through controlled deployment benchmarks. Third, the framework combines multi-tier explainability through SHAP feature attributions, graph influence analysis, and LLM-generated plain-language rationales. This contribution helps support transparent decision-making and traceable case review.

While broader adoption requires channel-specific validation on governed real data, as detailed in the next section, our multi-view ensemble ML framework provides a potentially transferable foundation for explainable operational fraud detection in other financial service channels.

\subsection{Limitations and Future Work}

The proposed framework exhibits several limitations that warrant consideration for production deployment and future research.

\paragraph{Synthetic Data Limitation.}
The training dataset is generated via rule-based probabilistic simulation with constraint enforcement and controlled fraud injection. While this approach preserves schema validity, regulatory compliance, and operational realism, some data generation rules may still be learnable by tree-based models. Consequently, absolute performance metrics should not be interpreted as production calibration. The ensemble's true discriminative performance requires validation on governed real data, where fraud patterns exhibit latent dependencies, human behavioral drift, and institution-specific edge cases that synthetic data cannot reproduce. As a next step, threshold tuning and policy calibration should be performed through retrospective validation on confirmed real-world outcomes.

\paragraph{Label Quality Limitation.}
The model assumes ground truth fraud labels are accurate and complete. In practice, fraud labels may rely on manual reporting, delayed discovery, and retrospective investigation, meaning some fraud cases remain unlabeled in training data. This label noise can bias the supervised component toward known fraud patterns and may underestimate risk for novel schemes not yet discovered and labeled. The unsupervised component may partially mitigate this limitation by detecting statistical outliers independent of labels, but overall ensemble calibration remains label-quality dependent. Future work can explore semi-supervised learning frameworks that leverage both labeled fraud cases and unlabeled suspicious patterns to improve detection coverage.

\paragraph{Evaluation Protocol Limitation.}
The train-test split is performed at the client level using stratified sampling to maintain fraud prevalence ratios, ensuring no client appears in both partitions. While this approach mitigates leakage, it does not preserve temporal ordering of fraud events. Because the split is random rather than sequential, training data may contain fraud cases that occurred chronologically after some test set cases. This violates realistic deployment scenarios where models are trained on historical data and evaluated on future fraud. A sequential train-test split—where all training fraud cases precede all test cases in time—would provide more rigorous evaluation of generalization to emerging fraud tactics. As a next step, future evaluation should explore time-based cross-validation strategies that partition data into sequential folds while maintaining sufficient fraud representation in each fold.

In addition, the stress test performed in Section~\ref{sec:experiment} focuses on a single shift dimension (account age) to simulate cold-start fraud risk in an elevated number of new clients. While this demonstrates robustness to one type of concept drift, real-world fraud landscapes exhibit multi-factor drifts that may interact in complex ways. Future work should test robustness under multi-dimensional drift scenarios, including: (1) burst of fraud cases concentrated in a particular industry sector, (2) shifting sender-network topology, and (3) seasonal transaction velocity spikes. Evaluating the ensemble under these compound shifts would provide stronger evidence of operational resilience.

\section{Acknowledgments}

We thank the RBC Amplify program for providing the opportunity to work on this project.

\bibliographystyle{plain}
\bibliography{reference}

\appendix

\section{Hyperparameter Configurations}
\label{appendix:hyperparameters}

This appendix provides the detailed hyperparameter configurations for all three constituent models in the ensemble. All hyperparameters were selected via Optuna optimization with 5-fold stratified cross-validation maximizing PR-AUC.

\subsection{XGBoost Hyperparameters}

Table \ref{tab:xgb_params} presents the optimal hyperparameters for the XGBoost fraud detection model, obtained from 50 Optuna trials with MedianPruner.

\begin{table}[h]
\centering
\caption{XGBoost Hyperparameter Configuration}
\label{tab:xgb_params}
\begin{tabular}{lll}
\hline
\textbf{Parameter} & \textbf{Value} & \textbf{Description} \\
\hline
\texttt{learning\_rate} & 0.163 & Step size shrinkage for boosting \\
\texttt{max\_depth} & 5 & Maximum tree depth \\
\texttt{subsample} & 0.951 & Fraction of samples per tree \\
\texttt{colsample\_bytree} & 0.560 & Fraction of features per tree \\
\texttt{scale\_pos\_weight} & 38.8 & Class imbalance weight (neg/pos ratio) \\
\texttt{objective} & binary:logistic & Binary classification with logistic output \\
\texttt{eval\_metric} & logloss & Training evaluation metric \\
\texttt{early\_stopping\_rounds} & 30 & Patience for early stopping \\
\texttt{random\_state} & 42 & Random seed for reproducibility \\
\hline
\end{tabular}
\end{table}

\subsection{Isolation Forest Hyperparameters}

Table \ref{tab:iso_params} presents the optimal hyperparameters for the Isolation Forest anomaly detection model, obtained from 50 Optuna trials with MedianPruner. Features are preprocessed via StandardScaler (mean=0, std=1) and SimpleImputer (median strategy), with correlation filtering applied at $|r| > 0.8$ threshold.

\begin{table}[h]
\centering
\caption{Isolation Forest Hyperparameter Configuration}
\label{tab:iso_params}
\begin{tabular}{lll}
\hline
\textbf{Parameter} & \textbf{Value} & \textbf{Description} \\
\hline
\texttt{n\_estimators} & 100 & Number of isolation trees \\
\texttt{max\_samples} & 0.795 & Fraction of samples per tree \\
\texttt{max\_features} & 0.495 & Fraction of features per tree \\
\texttt{contamination} & 0.03 & Expected fraud proportion (3\% threshold) \\
\texttt{n\_jobs} & -1 & Parallel processing (all cores) \\
\texttt{random\_state} & 42 & Random seed for reproducibility \\
\hline
\multicolumn{3}{l}{\textbf{Preprocessing}} \\
\hline
Feature scaling & StandardScaler & Mean=0, std=1 normalization \\
Missing values & SimpleImputer & Median imputation strategy \\
Correlation filter & $|r| > 0.8$ & Remove highly correlated features \\
\hline
\end{tabular}
\end{table}

\subsection{GraphSAGE Hyperparameters}

Table \ref{tab:gnn_params} presents the optimal hyperparameters for the GraphSAGE graph neural network model, obtained from 15 Optuna trials with MedianPruner. The graph comprises 89-dimensional node features (in/out aggregations of 10 transaction features + external indicator) and 10-dimensional edge features.

\begin{table}[h]
\centering
\caption{GraphSAGE Hyperparameter Configuration}
\label{tab:gnn_params}
\begin{tabular}{lll}
\hline
\textbf{Parameter} & \textbf{Value} & \textbf{Description} \\
\hline
\multicolumn{3}{l}{\textbf{Architecture}} \\
\hline
\texttt{hidden\_dim} & 160 & Hidden layer dimension \\
\texttt{embedding\_dim} & 48 & Final node embedding dimension \\
\texttt{num\_layers} & 2 & Number of GraphSAGE layers \\
\texttt{aggregator} & hybrid & Max + mean + sum concatenation \\
\texttt{activation} & ReLU & Non-linear activation function \\
\hline
\multicolumn{3}{l}{\textbf{Regularization}} \\
\hline
\texttt{dropout} & 0.42 & Dropout rate (conv layers + MLP) \\
\texttt{weight\_decay} & 0.0096 & L2 regularization coefficient \\
\hline
\multicolumn{3}{l}{\textbf{Training}} \\
\hline
\texttt{learning\_rate} & 0.0088 & Adam optimizer learning rate \\
\texttt{epochs} & 100 & Training iterations \\
\texttt{optimizer} & Adam & Adaptive moment estimation \\
\texttt{device} & CPU & PyTorch execution device \\
\hline
\multicolumn{3}{l}{\textbf{MLP Classification Head}} \\
\hline
Layer 1 & Linear(64 $\to$ 32) & Fully connected layer \\
Activation & ReLU & Non-linear activation \\
Regularization & Dropout(0.42) & Dropout layer \\
Layer 2 & Linear(32 $\to$ 1) & Output layer (fraud probability) \\
\hline
\multicolumn{3}{l}{\textbf{Graph Structure}} \\
\hline
Node features & 89-dim & In/out transaction aggregations + external flag \\
Edge features & 10-dim & Transaction statistics per edge \\
Total nodes & 25,093 & 17,093 clients + 8,000 external accounts \\
Total edges & 967,522 & Directed sender-receiver pairs \\
\hline
\end{tabular}
\end{table}

\section{Data Pipeline}
\label{appendix:data}
\subsection{Data Preparation.}
The dataset employed in this study is fully synthetic and was generated via a rule-based, probabilistic simulation pipeline designed to replicate the characteristics of after-hours business deposit service data. The dataset schema for deposits and cheques—including column definitions, data types, and relational constraints—was validated to ensure alignment with typical production systems. We first extracted canonical schemas and business constraints from deposit and fraud detection system data models, then generated new records by sampling from target distributions and enforcing domain-specific rules. The generation process follows three stages: (1) \textit{synthetic record generation}, where new entities (clients, owners, accounts, deposits, cheques, account transactions) are sampled to match expected operational volumes and behavioral patterns; (2) \textit{constraint-based simulation}, where cheque routing number formats, account relationships, timestamp ordering, currency compatibility, and transaction semantics are enforced during generation; and (3) \textit{fraud scenario injection}, where fraud labels (\texttt{is\_fraud}) are generated through controlled risk cohorts with overlapping anomaly segments to ensure fraud is not perfectly separable from legitimate behavior, thereby simulating real-world detection challenges.

The synthetic generation process preserves critical structural and domain properties while masking sensitive content. Preserved elements include source-system schema shapes and key joins (e.g., deposit IDs, client IDs, account numbers), domain mechanics (cheque routing structure, processing-time consistency, currency as an account property), and macro-level class balance (e.g., monthly seasonality, sender scenario proportions). Regenerated elements include all identifiers and personally identifiable information (PII)-like profile content, account numbers, routing strings, timestamps, IP patterns, transaction amounts, sender relationships, and fraud events.

Validation of the synthetic dataset is done as follows. Structural checks verify schema compliance, row counts, uniqueness constraints, and foreign key integrity. Behavioral checks enforce fraud-rate bounds, maximum fraud per client, timestamp ordering and windowing, currency integrity, transaction-type semantics, and sender-pool bounds. Stress and noise checks inject anomaly-fraud overlap, seasonal spikes, high-velocity bursts, dormant-then-burst behavior, and edge-case scenarios (e.g., a domestic currency account received cheques from foreign financial institutions) to test model robustness. The dataset is appropriate for end-to-end pipeline testing, feature engineering validation, model benchmarking under controlled class imbalance, and rule stress testing. However, synthetic data cannot reproduce all latent real-world dependencies, human behavioral drift, or institution-specific edge cases. Consequently, absolute performance metrics and parameters covered in this paper should not be treated as production calibration; final threshold tuning and policy decisions require evaluation on governed real data under operational conditions.

\subsection{Data scope}
The synthetic dataset comprises 17,093 after-hours business deposit client profiles, 480,000 after-hours business deposit slips, 2,000,000 after-hours business deposit cheques, and 1,500,000 business account transaction history entries spanning 12 months. These datasets represent linked entities within a unified relational data model, with magnitudes validated to reflect operational realities. The data model is structured hierarchically across four granularity levels: 
\begin{itemize}
    \item \textbf{Client Level}, where each business client is identified by a unique client ID and associated with one or more beneficial owners, with attributes including business type, jurisdiction, and deposit activity history; 
    \item \textbf{Business-Account Level}, where each client holds 1--5 accounts, each identified by a unique account number and associated with one or more owners, with account-specific attributes including currency (domestic or foreign) and transaction history; 
    \item \textbf{Deposit Level}, where each deposit, identified by a deposit slip, belongs to one depositing account and represents a single deposit event, with account number inferred from the deposit routing field;
    \item \textbf{Cheque Level}, where each cheque row belongs to exactly one deposit (deposit ID), establishing a one-to-many relationship between deposit and cheques. The distribution of cheques per deposit is intentionally skewed to match operational patterns: the mean is 4.2 cheques per deposit, but the tail extends to deposits containing 20+ cheques, reflecting bulk deposit behavior observed in commercial banking.
\end{itemize}

The distribution of dataset is tuned to simulate realism in financial institutions. The client population is dominated by domestic currency accounts, with a smaller foreign currency subset representing cross-border operations. Business sectors are sampled from a mix of categories, including restaurants, caterers, hotels, business services, auto dealers, software vendors, miscellaneous retail, and financial services and identified by Standard Industrial Classification (SIC) codes. Of all fraudulent clients, those with high-risk SIC codes account for approximately 35\% in our synthetic dataset.  Jurisdictional diversity is maintained through multiple domestic and cross-border sender-receiver patterns: cheques are sent from either the receiving institution, other domestic financial institutions, or foreign financial institutions. This multi-dimensional heterogeneity ensures that the dataset captures the complexity of real-world fraud detection, where fraudulent behavior is not confined to a single business sector, jurisdiction, or sender-receiver pattern.

\subsection{Feature Engineering}
We engineer a multi-view feature representation to capture complementary fraud signals across tabular and graph modalities. These features map out the following aspects:

\begin{itemize}
    \item \textbf{Business Profile}: 45 dimensions after one-hot encoding, including industry classification, account tenure and service enrollment flags;
    \item \textbf{Deposit Behavior}: 15 dimensions, including IP address deviations, transaction code or routing code mismatches, deposit outlier rates and normalized amounts;
    \item \textbf{Cheque History}: 20 dimensions, including cheque deposit velocity and amount rolling windows of 1 day, 7 days, 30 days and 90 days for domestic and foreign currencies separately;
    \item \textbf{Account Health}: 15 dimensions, including balance trends and volatility, activity patterns, dormancy and inbound/outbound flow anomalies;
    \item \textbf{Owner Profile}: 5 dimensions, including credit score, mortgage, loan, primary business account balances, total business account balances.
\end{itemize}

\end{document}